\documentclass[nohyperref]{article}

\usepackage{microtype}
\usepackage{graphicx}
\usepackage{subfigure}
\usepackage{booktabs} 

\usepackage{hyperref}

\usepackage[accepted]{icml2022}

\usepackage{amsmath}
\usepackage{amssymb}
\usepackage{mathtools}
\usepackage{amsthm}

\usepackage[capitalize,noabbrev]{cleveref}

\theoremstyle{plain}

\theoremstyle{definition}

\theoremstyle{remark}

\usepackage[textsize=tiny]{todonotes}

\icmltitlerunning{BEAR
for Learning Unbiased C2C Image Representations}

\usepackage{tikz}
\usetikzlibrary{backgrounds}
\usetikzlibrary{arrows,shapes}
\usetikzlibrary{tikzmark}
\usetikzlibrary{calc}

\usepackage{nccmath}
\usepackage{wrapfig}
\usepackage{comment}

\usepackage{blindtext}

\usepackage{xspace}

\usepackage{array}
\usepackage{ragged2e}
\newcolumntype{P}[1]{>{\RaggedRight\hspace{0pt}}p{#1}}
\newcolumntype{X}[1]{>{\RaggedRight\hspace*{0pt}}p{#1}}

\usepackage{tcolorbox}

\usepackage{tikz}
\usetikzlibrary{arrows,shapes,positioning,shadows,trees,mindmap}
\usepackage[edges]{forest}
\usetikzlibrary{arrows.meta}
\colorlet{linecol}{black!75}
\usepackage{xkcdcolors} 

\usepackage{tikz}
\usetikzlibrary{backgrounds}
\usetikzlibrary{arrows,shapes}
\usetikzlibrary{tikzmark}
\usetikzlibrary{calc}
\newcommand{\highlight}[2]{\colorbox{#1!17}{$\displaystyle #2$}}

\colorlet{mhpurple}{Plum!80}

\renewcommand{\highlight}[2]{\colorbox{#1!17}{#2}}

\begin{document}

\twocolumn[
\icmltitle{Bottleneck-based Encoder-decoder ARchitecture (BEAR)\\for Learning Unbiased Consumer-to-Consumer Image Representations}



\icmlsetsymbol{equal}{*}

\begin{icmlauthorlist}
\icmlauthor{Pablo Rivas}{bcs}
\icmlauthor{Gisela Bichler}{equal,csu}
\icmlauthor{Tomas Cerny}{equal,bcs}
\icmlauthor{Laurie Giddens}{equal,unt}
\icmlauthor{Stacie Petter}{equal,bis}
\end{icmlauthorlist}

\icmlaffiliation{bcs}{Department of Computer Science, Baylor University, Texas, USA}
\icmlaffiliation{csu}{School of Criminology and Criminal Justice, California State University, San Bernardino, California, USA}
\icmlaffiliation{unt}{Information Technology and Decisions Sciences Department, University of North Texas, USA}
\icmlaffiliation{bis}{School of Business, Wake Forest University, North Carolina, USA}

\icmlcorrespondingauthor{Pablo Rivas}{Pablo\_Rivas@Baylor.edu}

\icmlkeywords{Machine Learning, ICML}

\vskip 0.3in
]



\printAffiliationsAndNotice{\icmlEqualContribution} 

\begin{abstract}
Unbiased representation learning is still an object of study under specific applications and contexts. Novel architectures are usually crafted to resolve particular problems using mixtures of fundamental pieces. This paper presents different image feature extraction mechanisms that work together with residual connections to encode perceptual image information in an autoencoder configuration. We use image data that aims to support a larger research agenda dealing with issues regarding criminal activity in consumer-to-consumer online platforms. Preliminary results suggest that the proposed architecture can learn rich spaces using ours and other image datasets resolving important challenges that are identified.
\end{abstract}

\section{Introduction}

Online consumer-to-consumer (C2C) transactions surged throughout the COVID-19 pandemic as individuals looked for opportunities to generate income by selling goods and services or sought to save money by purchasing secondhand items \cite{goddevrind2021c2c}. Unfortunately, some online C2C marketplaces are vulnerable to criminal exploitation in which stolen or trafficked goods and services are sold. To sell stolen or trafficked goods and services in online C2C marketplaces, criminals must victimize individuals by stealing physical items or forcing victims to engage in services on behalf of the trafficker. When consumers discover they have purchased illicit goods or services in online C2C marketplaces, consumers' trust in these markets diminishes. Stakeholders and law enforcement agencies are combating the explosive growth in illicit e-commerce by conducting large-scale investigations  \cite{ballhaus2021ben,cbs2020major}. Scientists are re-calibrating explanations of criminal behavior and modifying crime prevention strategies in the face of dynamic post-COVID crime opportunity structures \cite{stickle2020crime}. 

\begin{figure}[t]
  \centering
   \includegraphics[width=\columnwidth]{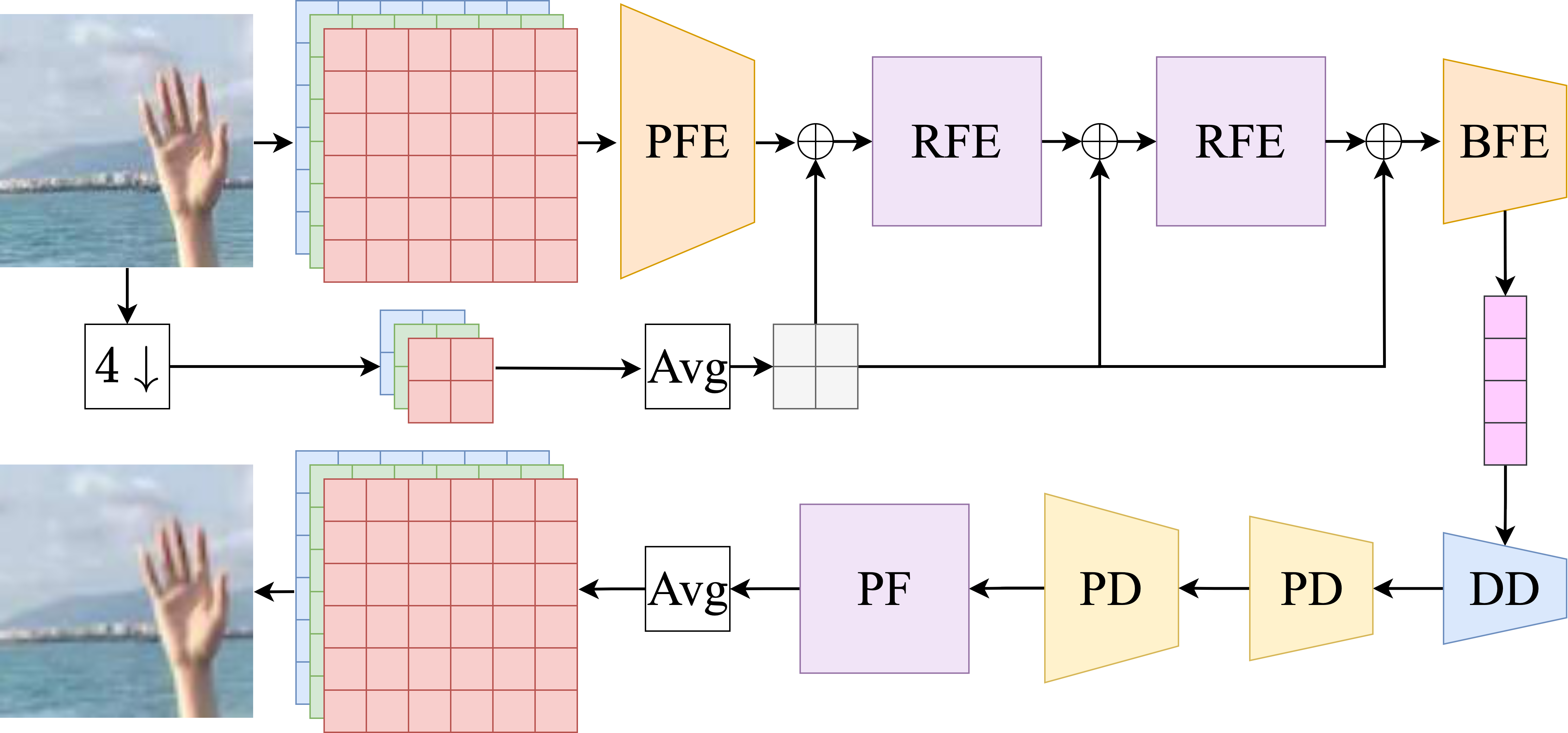}

   \caption{Proposed residual autoencoder that uses convolutional LSTMs for perceptual information extraction and compression.}
   \label{fig:arch}
\end{figure}

Stakeholders and scholars are searching for commonalities in text, and image-based expressions that may signal a potential victim of sex trafficking \cite{dubrawski2015leveraging,ibanez2014detection} to develop efficient deep learning-based models, such as the one we introduce in Figure \ref{fig:arch}, to aid detection efforts for identifying online advertisements that likely involve sex trafficking victims \cite{kejriwal2017flagit,thorn,tong2017combating}. Such efforts would have an even greater impact on crime reduction if machine learning algorithms could identify universal trafficking properties.

Currently, there are insufficient resources to monitor all online platforms and advertisements for suspicious activity. This leads our team to inquire how machine learning-based NLP and computer vision can be used to find markers and patterns of suspected online illicit activity and whether such models generalize to different commodities and marketplaces. While this is a large multidisciplinary project that will span two years, this paper focuses on one of the initial machine learning challenges, i.e., the encoding of C2C images that will enable independent image analysis and multimodal architecture design that integrates text as part of downstream trafficking detection tasks. Specifically, the challenges dealing with images from C2C marketplaces related to trafficking are many, including: i) that investigators should minimize exposure to pornographic material; ii) the amount of image data posted daily on these platforms is impossible to analyze manually by humans; iii) that although the data is publicly available, it may contain biometric identifiers (e.g., face) and other personal identity markers readily available in its native image format, but should not. Therefore, our approach to these challenges is reflected in our contributions, which are as follows:
\begin{enumerate}
  \item we design and develop an autoencoder model that will learn rich features that will produce a latent representation of an image;
  \item our model uses modern machine learning techniques to learn low-dimensional representations at scale with few parameters in comparison to other similar models;
  \item the proposed model hides personal identifiers from plain sight while keeping necessary information for downstream trafficking detection tasks.
\end{enumerate} 

While other important methodologies such as Vision Transformers \cite{dosovitskiy2020image}, BYOL \cite{grill2020bootstrap}, or CLIP \cite{radford2021learning} represent great advantages in the field, they also bring some undesirable effects into our research, which we discuss next in Section 2 along with a brief background on autoencoders. Section 3 introduces the proposed architecture; Section 4 discusses our results, and we conclude in Section 5.

\section{Background}

In machine learning, researchers are particularly interested in exploring different learning paradigms, including supervised and unsupervised, as well as other recent variants that include self-supervised, contrastive learning, etc. For our particular problem, some of these paradigms are not immediately applicable; our problem statement is to find an architecture and learning paradigm for image representations with a lightweight, scalable model that hides from plain sight personal identifiers, yet it needs to embed sufficient information about image perception for downstream multimodal human trafficking tasks.

\citet{dosovitskiy2020image} recently introduced the \emph{Vision Transformer} (ViT), which is an attention-based encoder mechanism that shows superior performance in downstream classification tasks. ViT demonstrated state-of-the-art performance in traditional image datasets for classification. Although the ViT is scalable and generalizes well, its major disadvantages are that it can be costly to train, it is data-hungry, and the baseline ViT, as is typical of most transformer architectures, has a large number of parameters; it can have between 86M and 632M of parameters. 

\citet{grill2020bootstrap} introduced the \emph{Bootstrap Your Own Latent} (BYOL) approach that uses a self-supervised approach to learn image representations based on sampling the distribution of the input image $\mathbf{x}$ on two networks, one that yields predictions and one that yields projections. The model then is optimized to minimize the loss:\\

\begin{equation}
    \mathcal{L}(\tikzmarknode{u}{\highlight{red}{$\theta,\xi$}}; \tikzmarknode{b}{\highlight{orange}{$\mathbf{z}_\theta,\mathbf{z}_\xi^\prime$}})=2-2 \cdot \frac{\left\langle q_{\theta}\left(\mathbf{z}_{\theta}\right), \mathbf{z}_{\xi}^{\prime}\right\rangle}{\left\|q_{\theta}\left(\mathbf{z}_{\theta}\right)\right\|_{2} \cdot\left\|\mathbf{z}_{\xi}^{\prime}\right\|_{2}} \nonumber
\end{equation}
\begin{tikzpicture}[overlay,remember picture,>=stealth,nodes={align=left,inner ysep=1pt},<-]
    \path (u.north) ++ (0,2em) node[anchor=south west,color=red!67] (scalep){\textbf{model model hyperparameters}};
    \draw [color=red!57](u.north) |- ([xshift=-0.3ex,color=red]scalep.south east);
    \path (b.south) ++ (0,-1.5em) node[anchor=north west,color=orange!67] (mean){\textbf{projections of latent representations}};
    \draw [color=orange!57](b.south) |- ([xshift=-0.3ex,color=orange]mean.south east);
\end{tikzpicture}

where $q_\theta(\cdot)$ is a linear predictor. This optimization ensures that the learned representations on the target network are not trivially predicted and promotes robustness in learning the distribution of the input at the level of the representation space. However, for our context, there are also disadvantages, i.e., the methodology relies on data for classification tasks or other label-based data, e.g., ImageNet, to guarantee sufficient data for bootstrapping and distribution approximation. Further, the baseline model, i.e., based on ResNets, is significantly large, from 250M to 375M of parameters. 

More recently, \citet{radford2021learning} made significant progress in contrastive learning by introducing a \emph{Contrastive Language–Image Pre-training} (CLIP) methodology. CLIP is a multimodal approach that considers language and image information to align the representation space of text and images that belong together, allowing for the navigation of the latent space either with text or images with the right prompts. Due to the nature of the language portion of CLIP, it can also be extremely costly to fully train CLIP on C2C text and image data. Therefore, in our context CLIP is more suitable for transfer learning from a pre-trained model on data similar to C2C marketplaces, e.g., social media. However, at this stage of our project, we are investigating only image representations that will obfuscate personally identifiable information, and minimize the investigator's exposure to explicit trafficking content, all the while using a lightweight model that still produces good latent representations. With this in mind, we propose to use an autoencoder architecture that learns convolutional filters, similar to ViT's patches, and that uses a similarity-based loss and two slightly different parallel networks as in BYOL. 

A convolutional autoencoder (CAE) is a type of architecture that falls within the CNN family, and it is traditionally trained using an unsupervised learning approach \cite{kumar_how_2020}. The most common use for CAEs is to learn low-dimensional image representations by minimizing an image reconstruction error criteria through learning optimal convolutional filters at different layers \cite{fukushima1982neocognitron}. CNNs offer additional intriguing characteristics and architectures that have resulted in new applications and variants \cite{aloysius_review_2017,sultana_review_2020,zhang_graph_2019}. Although many AEs, such as the variational autoencoder (VAE) and the denoising AE, have been successfully implemented \cite{10.1145/3106426.3106489,10.1145/3234944.3234956,10.1145/3126686.3126774,wetzel_unsupervised_2017,10.1145/3388440.3412458}, there is still research being done in alternative ways to make them more efficient during learning, or making them optimize and generalize better. 

The proposed architecture can be similar to the work by \citet{zhou2020salient} which uses residual connections in an image autoencoder; and the work of \citet{zhao2021combination} that takes advantage of temporal data. However, our work is unique in using several elements together to produce rich representations that solve the problem defined earlier.

\section{Experimental Architecture}

As established earlier, our BEAR autoencoder is composed of different light-weight feature extraction mechanisms that take advantage of spatial relationships and residual information across layers. A square input image $\mathbf{x} \in \mathbb{R}^{n \times n \times d}$ with $n$ rows, $n$ columns, and an arbitrary depth $d$, in our context is a classic three-channel color image. The input image is also downsampled by a factor of $4$ in all dimensions except depth, $\mathbf{\tilde{x}} \in \mathbb{R}^{n/4 \times n/4 \times d}$. While $\mathbf{x}$ feeds the main feature extraction pipeline, $\mathbf{\tilde{x}}$ feeds the residual connections.

\begin{figure*}[h]
  \centering
   \includegraphics[width=0.65\textwidth]{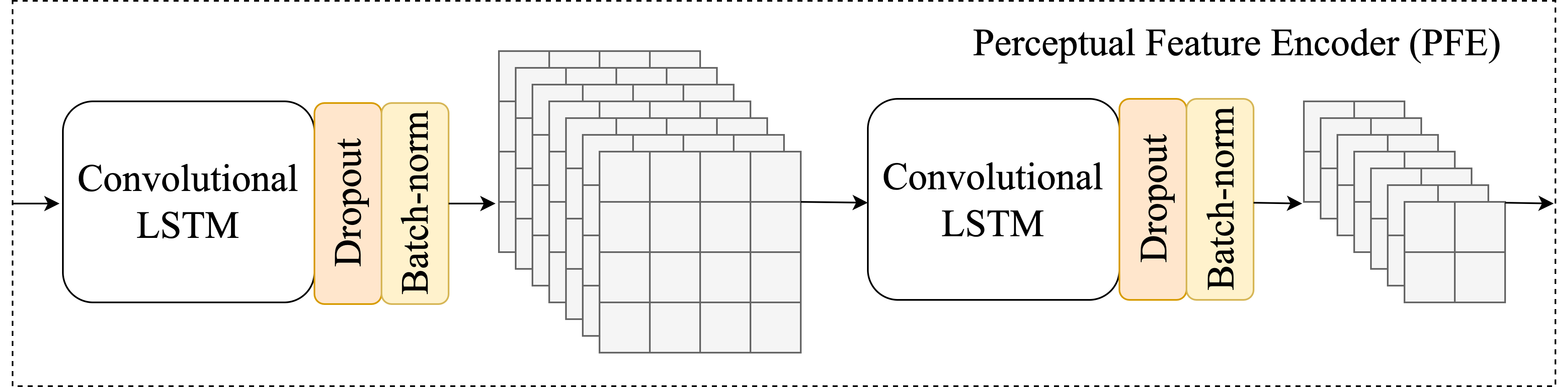}

   \caption{The \emph{Perceptual Feature Encoder} architecture that reduces dimensions while extracting perceptual information.}
   \label{fig:PFE}
\end{figure*}

\begin{figure*}[h]
  \centering
   \includegraphics[width=0.6\textwidth]{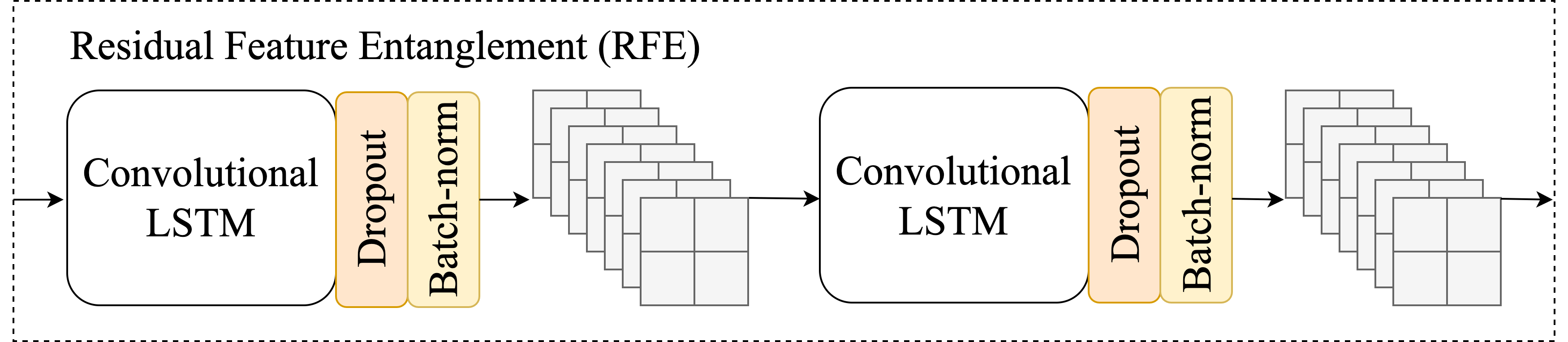}

   \caption{The \emph{Residual Feature Entanglement} that uses as input both the previous layer information and the residual, preserving dimensions.}
   \label{fig:RFE}
\end{figure*}

\begin{figure}[h]
  \centering
   \includegraphics[width=0.8\columnwidth]{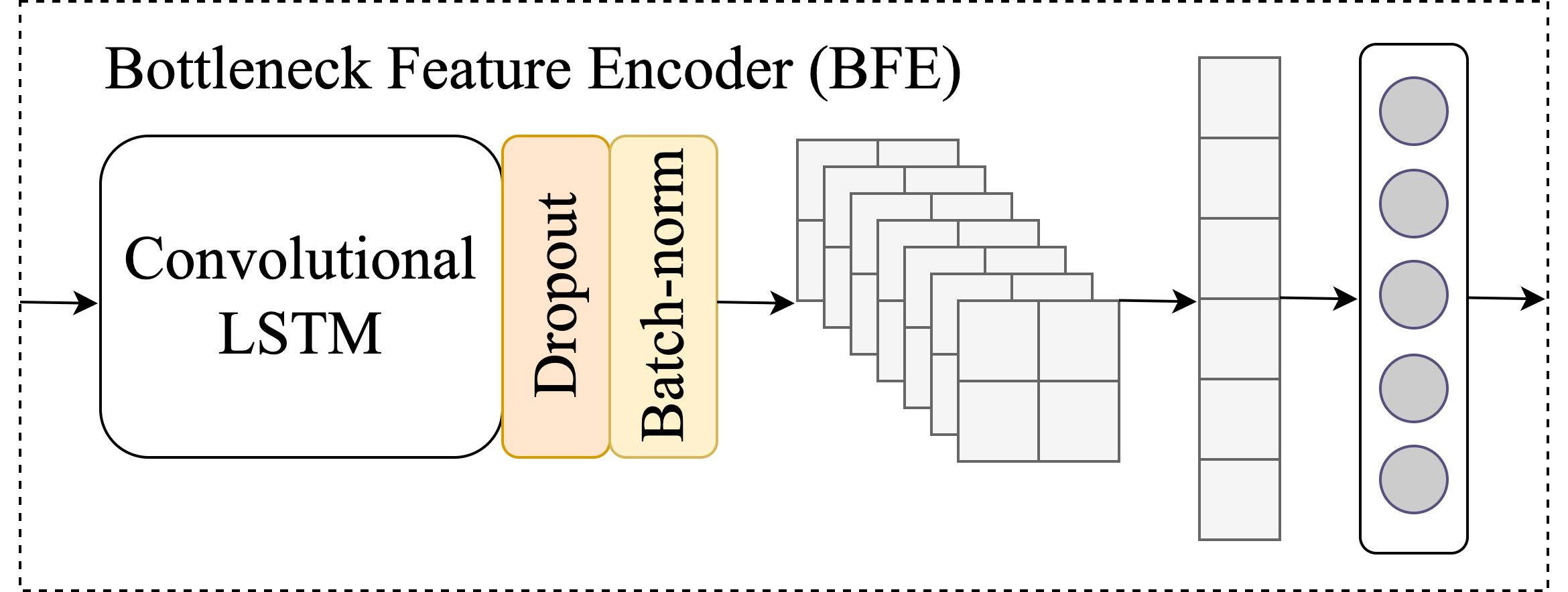}

   \caption{The BFE uses a convolutional LSTM and a dense layer.}
   \label{fig:BFE}
\end{figure}

\subsection{Encoder}
From Figure \ref{fig:arch} we can observe that the first piece of the BEAR model goes through a \emph{Perceptual Feature Encoder}, $\mathbf{z}_\text{PFE}=\text{PFE}(\mathbf{x};\theta)$, which is composed of two blocks of regularized convolutional long-short term memory models (LSTMs), first introduced by \citet{shi2015convolutional}, which yield features $\mathbf{z}_\text{PFE}$, see Figure \ref{fig:PFE}. Then the data goes through a \emph{Residual Feature Entanglement} piece, $\mathbf{z}_\text{RFE}=\text{RFE}(\mathbf{z}_\text{PFE},\mathbf{\tilde{x}};\theta)$, which uses perceptual features and the residual from the original downsampled image; this process is done twice back to back, see Figure \ref{fig:RFE}. Note that the RFE segment ought to be removed if anyone fine-tunes the model for downstream object recognition tasks, as feature entanglement is know to have detrimental effects in fitting labels \cite{zheng2021improving}; however, feature entanglement is naturally beneficial for unsupervised image representation learning such as in our context; both of these go into the encoder, as we discuss next. 

Next the model goes into a \emph{Bottleneck Feature Encoder}, $\mathbf{z}_\text{BFE}=\text{BFE}(\mathbf{z}_\text{RFE},\mathbf{\tilde{x}};\theta)$, which is comprised of a convolutional LSTM and a dense layer, see Figure \ref{fig:BFE}. Both PFE and BFE reduce dimensions as part of the process, and BFE is the last piece of the encoding process. Thus, $\mathbf{z}_\text{BFE} \in \mathbb{R}^m$ is the learned representation of the input in a lower $m-$dimensional space, i.e., $\mathbb{R}^{256}$.

\begin{figure}[h!]
  \centering
   \includegraphics[width=0.65\columnwidth]{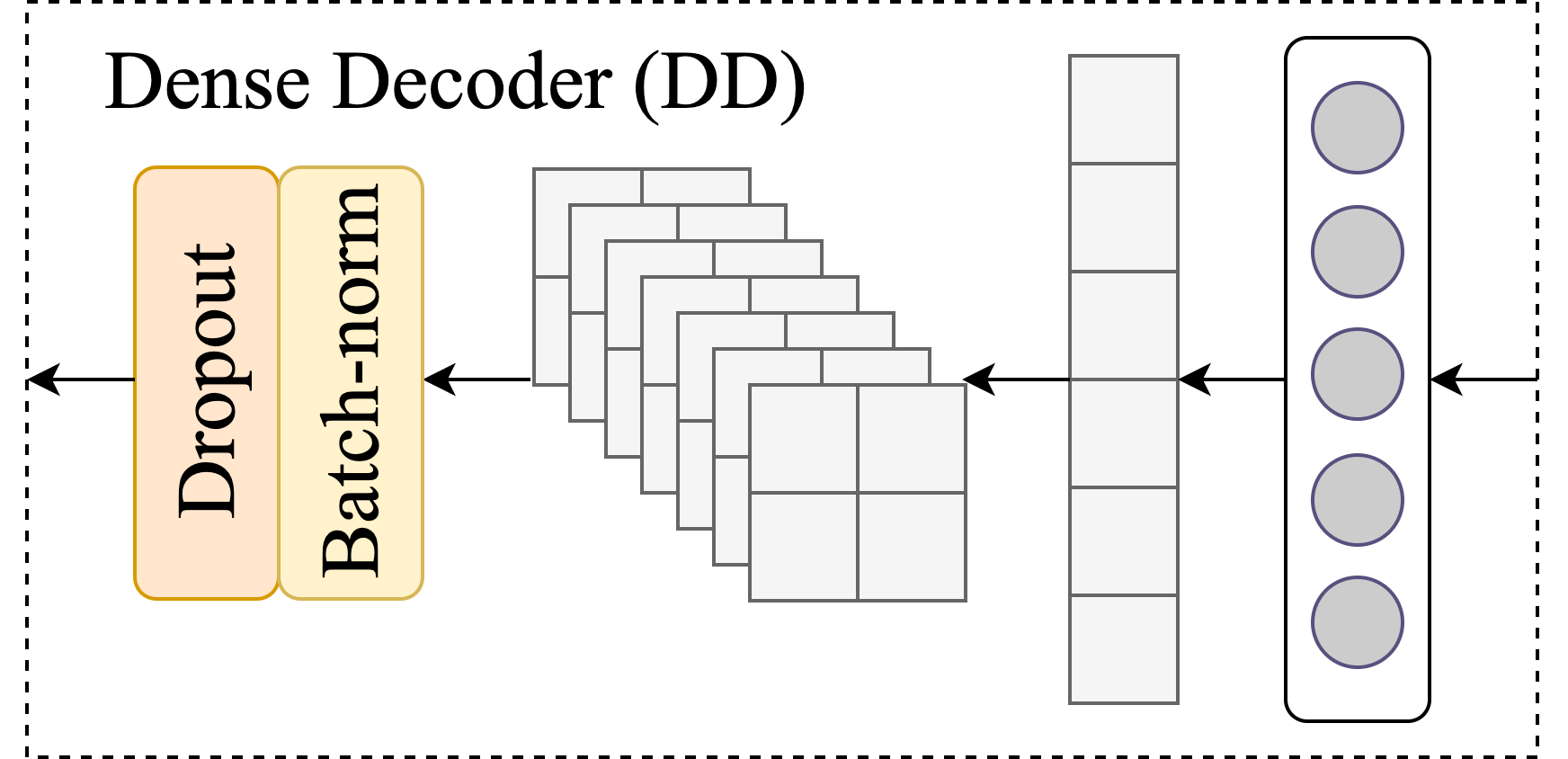}

   \caption{The DE reconstructs feature maps using a dense layer.}
   \label{fig:DD}
\end{figure}

\begin{figure}[h]
  \centering
   \includegraphics[width=\columnwidth]{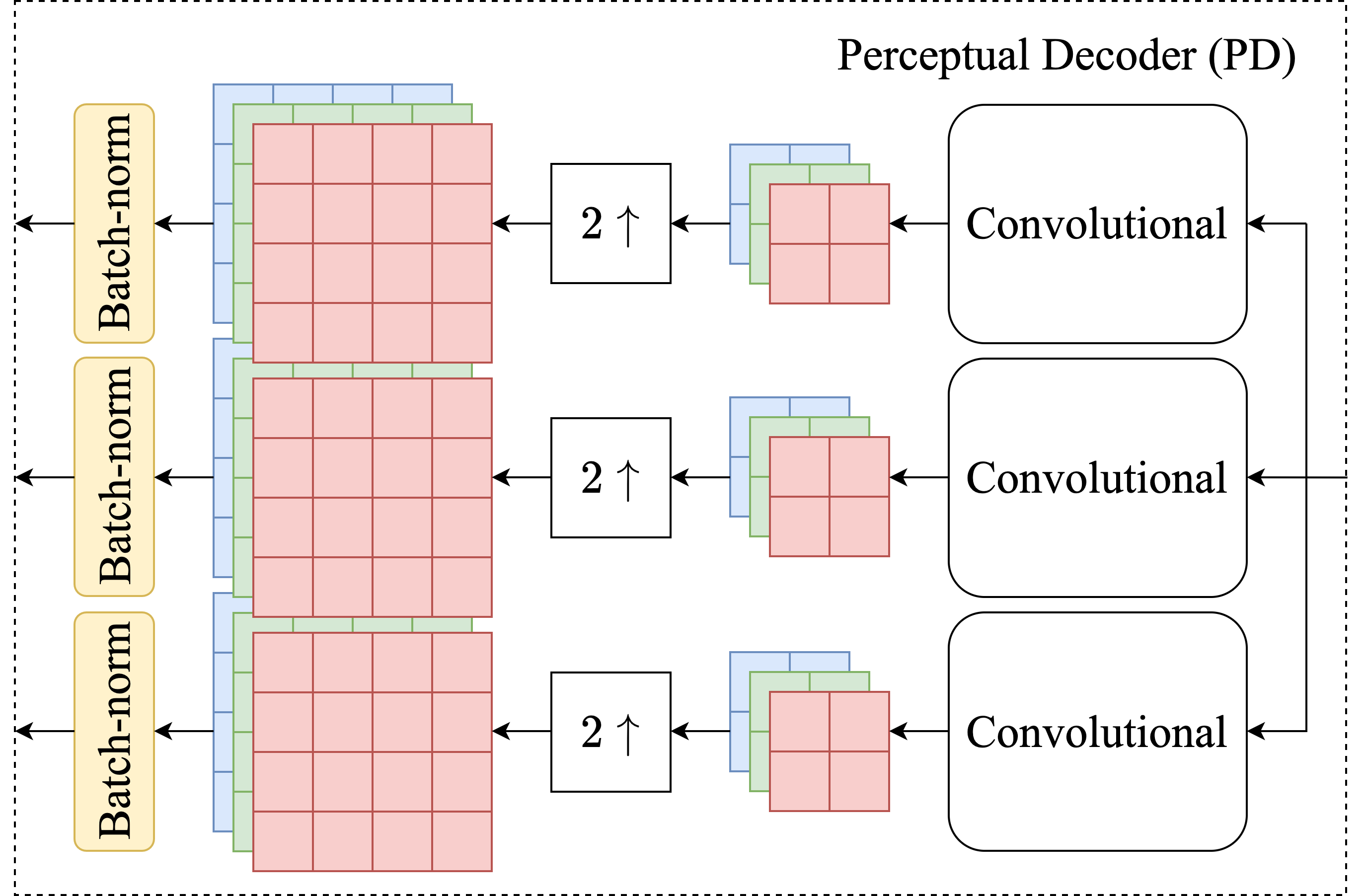}

   \caption{The PD uses parallel convolutions before up-sampling.}
   \label{fig:PD}
\end{figure}

\begin{figure}[h]
  \centering
   \includegraphics[width=0.8\columnwidth]{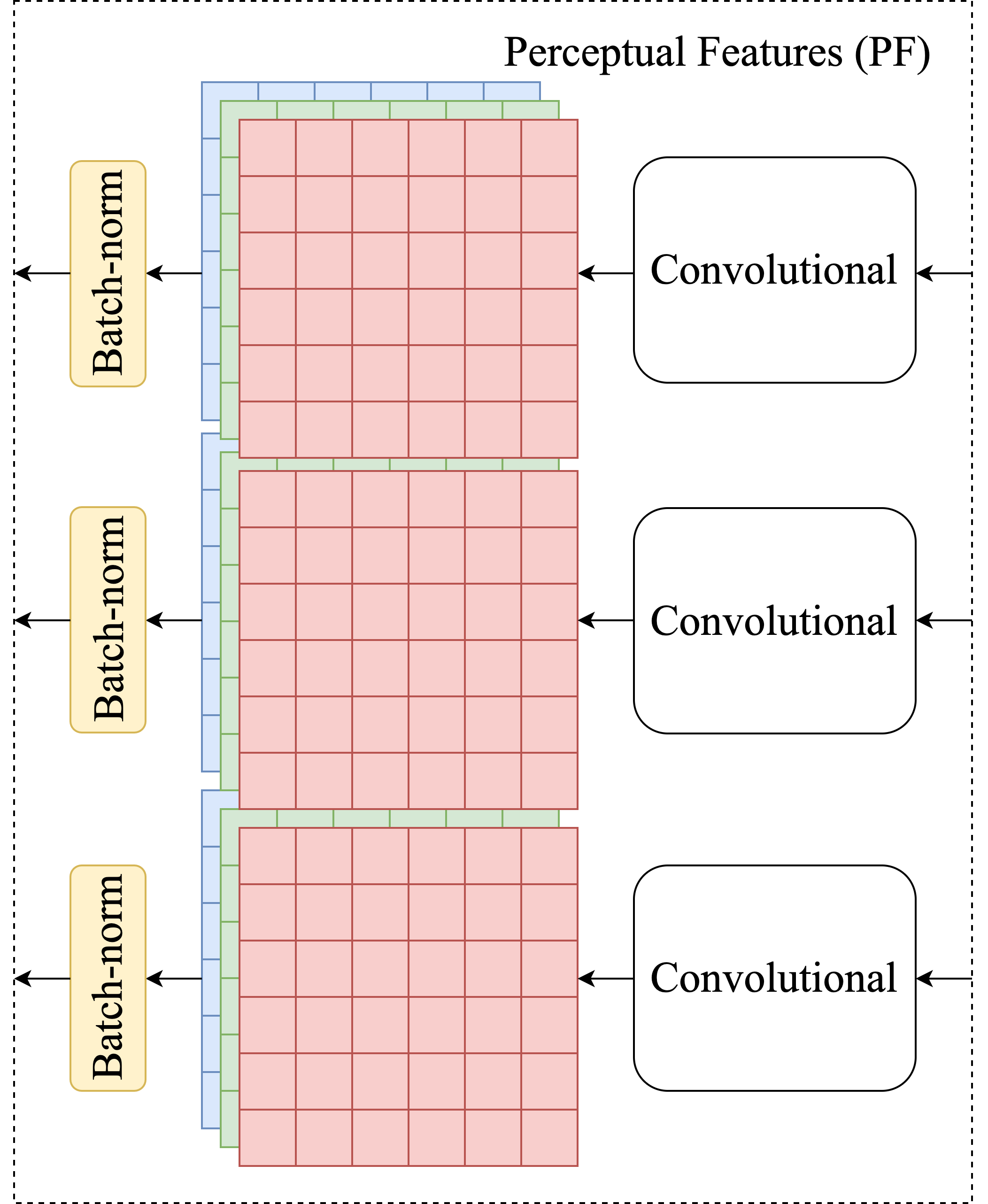}

   \caption{The PF reconstructs perceptual features with parallel convolutions.}
   \label{fig:PF}
\end{figure}

\subsection{Decoder}
The decoder in the architecture is relatively different in the sense that it has no residual connections and starts with a \emph{Dense Decoder}, $\mathbf{z}_\text{DD}=\text{DD}(\mathbf{z}_\text{BFE};\theta)$, which begins with a dense layer that produces features to facilitate reconstruction, see Figure \ref{fig:DD}. Next, we reconstruct increasing dimensions through back-to-back \emph{Perceptual Decoders}, $\mathbf{z}_\text{PD}=\text{PD}(\mathbf{z}_\text{DD};\theta)$, and $\mathbf{z}_\text{PD}=\text{PD}(\mathbf{z}_\text{PD};\theta)$, both of which are based on parallel convolutional networks, see Figure \ref{fig:PD}. Finally, we have trainable \emph{Perceptual Features}, $\mathbf{z}_\text{PF}=\text{PF}(\mathbf{z}_\text{PD};\theta)$, consisting on parallel convolutional networks for larger feature re-composition, see Figure \ref{fig:PF}. The reconstructed image is obtained by averaging the channels of the PF, $\mathbf{\hat{x}}=\text{Avg}(\mathbf{z}_\text{PF})$. 
\subsection{Training}

The BEAR model is trained using a classic binary cross entropy loss given by:\\

\begin{equation}
\label{eq:loss}
    \mathcal{L}(\tikzmarknode{u}{\highlight{red}{$\theta$}}; \tikzmarknode{b}{\highlight{blue}{$\mathbf{x}$}})=-\frac{1}{d} \sum_{i=1}^{d} \mathbf{x}_{i} \cdot \log \hat{\mathbf{x}}_{i}+\left(1-\mathbf{x}_{i}\right) \cdot \log \left(1-\hat{\mathbf{x}}_{i}\right)
\end{equation}
\begin{tikzpicture}[overlay,remember picture,>=stealth,nodes={align=left,inner ysep=1pt},<-]
    \path (u.north) ++ (0,2em) node[anchor=south west,color=red!67] (scalep){\textbf{autoencoder model hyperparameters}};
    \draw [color=red!57](u.north) |- ([xshift=-0.3ex,color=red]scalep.south east);
    \path (b.south) ++ (0,-1.5em) node[anchor=north west,color=blue!67] (mean){\textbf{input image in $\mathbb{R}^{n \times n \times d}$}};
    \draw [color=blue!57](b.south) |- ([xshift=-0.3ex,color=blue]mean.south east);
\end{tikzpicture}

where $\mathbf{x} \in \mathbb{R}^{n \times n \times d}$ is the input, $\mathbf{\hat{x}}$ is the reconstruction, and $\theta$ are the model parameters. This loss can also be replaced with an MSE loss achieving similar results. Minimizing the loss in (\ref{eq:loss}) repeatedly, using random starts, yields consistent results across the board using the Adam optimizer with an initial learning rate of $0.0001$ and an automatic decay after a five epoch plateau. The learning stops after no improvements in the validation loss after 10 epochs.

The dataset consists of images of C2C sites publicly available. However, due to privacy concerns, our data is not publicly available to protect the potentially identifiable private information of humans. The total number of images is 1,998,680; they all vary in size originally but are resized to $128 \times 128 \times 3$ during training. The results obtained are discussed next.

\section{Results}

We trained the proposed architecture, as discussed in the previous section, using different three data sets as indicated in Table \ref{tbl:data}. The first dataset is C2C image data publicly available on the web; however, we do not possess the rights to distribute. This dataset contains images of people, objects, text, numbers, social media accounts, etc. These may contain images of trafficking victims although the images are not labeled, since this is a future task for a different stage of our project. Our intention on using the C2C data at this early stage of our entire research project is to test the ability of our model to achieve fast converge and fast throughput, while producing image representations that obfuscate personal identifiers and information that is readily available visually. 

\begin{table}[t]
\caption{Data set comparison with respect to convergence time.}
\label{tbl:data}
\vskip 0.15in
\begin{center}
\begin{small}
\begin{sc}
\begin{tabular}{lcccr}
\toprule
Data set & $N$ & Size & Labeled & Time (h)\\
\midrule
C2C    & 1.9m & varies & $\times$ & 132 \\
CIFAR-10 & 60k & $32 \times 32$ & $\surd$ & 20 \\
ImageNet    & 1.2m & varies & $\surd$ & 99 \\
\bottomrule
\end{tabular}
\end{sc}
\end{small}
\end{center}
\vskip -0.1in
\end{table}
 
The experiment in Table \ref{tbl:data} is to illustrate the convergence time required to fully train the BEAR architecture on C2C data and two other popular datasets. The table gives an idea of the training time on a single GPU NVIDIA 1080 card with 8GB. This suggests that the model behaves consistently with respect to the size of the dataset. 

\begin{figure}[t]
  \centering
   \includegraphics[width=\columnwidth]{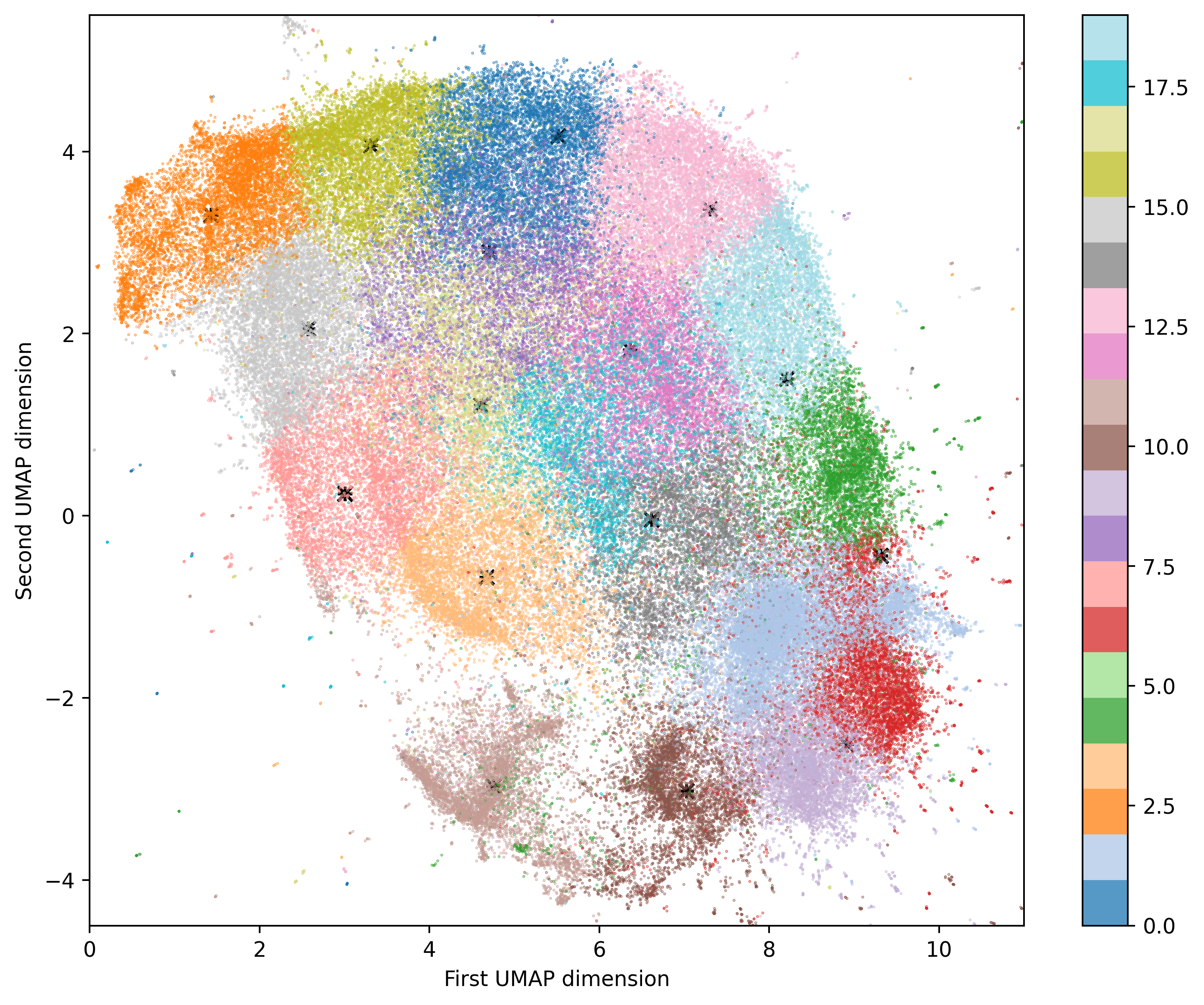}

   \caption{Learned representations on C2C data visualized using a 2D~UMAP and $k-$means with $k=20$.}
   \label{fig:bear-umap}
\end{figure}

\begin{figure}[h!]
  \centering
   \includegraphics[width=\columnwidth]{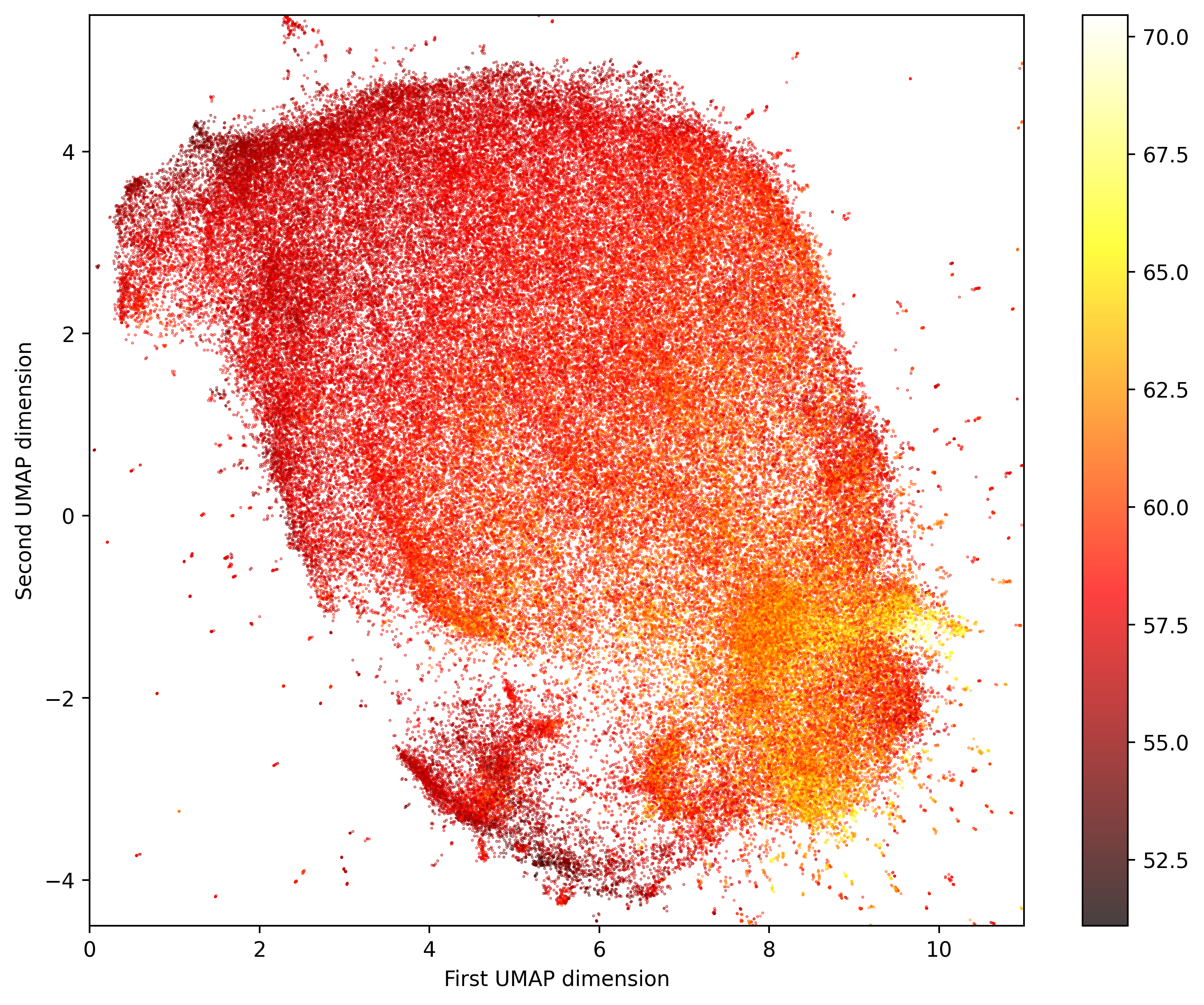}

   \caption{Learned representations on C2C data visualized using $||\mathbf{z}_\text{BFE}||_2$ as color reference.}
   \label{fig:bear-umap-norm}
\end{figure}

Next, we are interested in analyzing the learned space and the representations, particularly in the context of C2C data. Figure \ref{fig:bear-umap} depicts the UMAP 2D representation of the learned representations using $k-$means to identify potential clusters \cite{becht2019dimensionality}. Using the elbow method, we identified $k$ between 10 and 20 to be a good number of clusters. The figure identifies 20 clusters with different colors and displays the centroids. The figure suggests that the latent space is formed of several distinct groups that are distributed across the space and clustered according to their perceptual information. Further examination in C2C data can demonstrate that cluster prototypes are distinct among each other, with respect to their visual characteristics beyond the trivial color information. 

To show further that the latent space distributes data displaying different information, we can simply look at the norm of the representation vector $||\mathbf{z}_\text{BFE}||_2$, as shown in Figure \ref{fig:bear-umap-norm}. From the figure, we can appreciate that the data points are often associated and grouped along with other vectors with similar perceptual information; however, the space is not limited or distributed solely by norm similarity. The feature space has perceptual image entanglement, which is a desired property prior to our future work, which includes re-training the model on labeled C2C data for detecting criminal activity in a multimodal setting using constrastive learning. 

Finally, we conducted another experiment consisting of running $k-$means not in a reduced 2D space but in a slightly higher-dimensional space, i.e., 20 dimensions, in comparison to the 10-dimensional space of Figure \ref{fig:bear-umap}, where we would obtain a result with slightly fuzzier cluster limits. These effects are usually expected due to the curse of dimensionality that affects $k-$means. The experiments were conducted in C2C, CIFAR-10, and ImageNet data. From these experiments, we can consistently observe that the data is distributed by perceptual information and not just by color codes; further, $k$-means converges consistently to satisfy the cluster quality objective: $\underset{\mathbf{S}}{\arg \min } \sum_{i=1}^{k} \sum_{\mathbf{z}_\text{BFE} \in S_{i}}\left\|\mathbf{z}_\text{BFE}-\boldsymbol{\mu}_{i}\right\|^{2}$, for partition sets $\mathbf{S}=\left\{S_{1}, S_{2}, \ldots, S_{k}\right\}$. This empirically shows that the model will be consistent across different image datasets.

In summary, the proposed model demonstrates the following. First, BEAR is able to converge to a local minima that satisfies a reconstruction loss, suggesting that the perceptual features extracted using light-weight network segments compresses sufficient information for other downstream tasks. Second, the model consistently performs in a stable manner during unsupervised learning for other popular image datasets. Third, BEAR can consistently satisfy a $k$-means cluster quality objective, suggesting that the latent space can induce regular groups naturally. Fourth, the model can hide from plain sight private personal information due to the natural abilities of autoencoders to compress data using unique transformations, as opposed to fixed compression methodologies based, for example, in basis functions from the discrete cosine transform (DCT), that can be used to recover DCT compressed data.

\section{Conclusions}

This paper presents BEAR, an autoencoder-based architecture that treats the channel information in images as sequential data in convolutional LSTMs and using residual connections to preserve features across different encoding pathways. The encoding-decoding architecture called BEAR was trained on C2C data aiming to model content that is typical of these platforms with the ultimate goal of learning robust image representations with a light-weight model. 

While other state-of-the-art image feature extraction methods such as VGG and Inception-based are powerful, they are usually biased toward specific classes through label-based supervised learning or semi-supervised for transfer learning; the inductive bias of these architectures favors a limited number of classification tasks. The proposed BEAR methodology is not particularly biased toward specific labels, but remains optimal in preserving perceptual features. 

In the future, we aim to use this architecture in connection with textual information and build a larger multimodal data pipeline that will aid in detecting online crime on C2C platforms using contrastive learning.

\section*{Acknowledgements}
This work was funded by the National Science Foundation under grants CHE-1905043, CNS-2136961, and CNS-2210091.


\bibliography{refs}

\begin{thebibliography}{27}
\providecommand{\natexlab}[1]{#1}
\providecommand{\url}[1]{\texttt{#1}}
\expandafter\ifx\csname urlstyle\endcsname\relax
  \providecommand{\doi}[1]{doi: #1}\else
  \providecommand{\doi}{doi: \begingroup \urlstyle{rm}\Url}\fi

\bibitem[Aloysius \& Geetha(2017)Aloysius and Geetha]{aloysius_review_2017}
Aloysius, N. and Geetha, M.
\newblock A review on deep convolutional neural networks.
\newblock \emph{2017 International Conference on Communication and Signal
  Processing (ICCSP)}, 2017.
\newblock \doi{10.1109/ICCSP.2017.8286426}.

\bibitem[Ballhaus \& Ramachandran(2021)Ballhaus and
  Ramachandran]{ballhaus2021ben}
Ballhaus, R. and Ramachandran, S.
\newblock Ben dugan works for cvs., Business. The Wall Street Journal.
  September 2021.
\newblock URL
  \url{https://www.wsj.com/articles/cvs-home-depot-theft-organized-crime}.

\bibitem[Becht et~al.(2019)Becht, McInnes, Healy, Dutertre, Kwok, Ng, Ginhoux,
  and Newell]{becht2019dimensionality}
Becht, E., McInnes, L., Healy, J., Dutertre, C.-A., Kwok, I.~W., Ng, L.~G.,
  Ginhoux, F., and Newell, E.~W.
\newblock Dimensionality reduction for visualizing single-cell data using umap.
\newblock \emph{Nature biotechnology}, 37\penalty0 (1):\penalty0 38--44, 2019.

\bibitem[Castellini et~al.(2017)Castellini, Poggioni, and
  Sorbi]{10.1145/3106426.3106489}
Castellini, J., Poggioni, V., and Sorbi, G.
\newblock Fake twitter followers detection by denoising autoencoder.
\newblock In \emph{Proceedings of the International Conference on Web
  Intelligence}, WI '17, pp.\  195–202, New York, NY, USA, 2017. Association
  for Computing Machinery.
\newblock ISBN 9781450349512.
\newblock \doi{10.1145/3106426.3106489}.
\newblock URL \url{https://doi.org/10.1145/3106426.3106489}.

\bibitem[CBS(2020)]{cbs2020major}
CBS, S.
\newblock Major san francisco bay area retail theft ring busted, crime section.
  san francisco, ca., October 2020.
\newblock URL \url{https://sanfrancisco.cbslocal.com/tag/stolen-property/}.

\bibitem[Dosovitskiy et~al.(2020)Dosovitskiy, Beyer, Kolesnikov, Weissenborn,
  Zhai, Unterthiner, Dehghani, Minderer, Heigold, Gelly,
  et~al.]{dosovitskiy2020image}
Dosovitskiy, A., Beyer, L., Kolesnikov, A., Weissenborn, D., Zhai, X.,
  Unterthiner, T., Dehghani, M., Minderer, M., Heigold, G., Gelly, S., et~al.
\newblock An image is worth 16x16 words: Transformers for image recognition at
  scale.
\newblock In \emph{International Conference on Learning Representations}, 2020.

\bibitem[Dubrawski et~al.(2015)Dubrawski, Miller, Barnes, Boecking, and
  Kennedy]{dubrawski2015leveraging}
Dubrawski, A., Miller, K., Barnes, M., Boecking, B., and Kennedy, E.
\newblock Leveraging publicly available data to discern patterns of
  human-trafficking activity.
\newblock \emph{Journal of Human Trafficking}, 1\penalty0 (1):\penalty0 65--85,
  2015.

\bibitem[Fukushima \& Miyake(1982)Fukushima and
  Miyake]{fukushima1982neocognitron}
Fukushima, K. and Miyake, S.
\newblock Neocognitron: A self-organizing neural network model for a mechanism
  of visual pattern recognition.
\newblock In \emph{Competition and cooperation in neural nets}, pp.\  267--285.
  Springer, 1982.

\bibitem[Goddevrind et~al.(2021)Goddevrind, Schumacher, Seetharaman, and
  Spillecke]{goddevrind2021c2c}
Goddevrind, V., Schumacher, T., Seetharaman, R., and Spillecke, D.
\newblock C2c e-commerce: Could a new business model sell more old goods?,
  McKinsey \& Company. Sep. 2021.
\newblock URL
  \url{https://www.mckinsey.com/industries/technology-media-and-telecommunications/our-insights/c2c-ecommerce-could-a-}.

\bibitem[Grill et~al.(2020)Grill, Strub, Altch{\'e}, Tallec, Richemond,
  Buchatskaya, Doersch, Avila~Pires, Guo, Gheshlaghi~Azar,
  et~al.]{grill2020bootstrap}
Grill, J.-B., Strub, F., Altch{\'e}, F., Tallec, C., Richemond, P.,
  Buchatskaya, E., Doersch, C., Avila~Pires, B., Guo, Z., Gheshlaghi~Azar, M.,
  et~al.
\newblock Bootstrap your own latent-a new approach to self-supervised learning.
\newblock \emph{Advances in Neural Information Processing Systems},
  33:\penalty0 21271--21284, 2020.

\bibitem[Ibanez \& Suthers(2014)Ibanez and Suthers]{ibanez2014detection}
Ibanez, M. and Suthers, D.~D.
\newblock Detection of domestic human trafficking indicators and movement
  trends using content available on open internet sources.
\newblock In \emph{2014 47th Hawaii international conference on system
  sciences}, pp.\  1556--1565. IEEE, 2014.

\bibitem[Jhamb et~al.(2018)Jhamb, Ebesu, and Fang]{10.1145/3234944.3234956}
Jhamb, Y., Ebesu, T., and Fang, Y.
\newblock Attentive contextual denoising autoencoder for recommendation.
\newblock In \emph{Proceedings of the 2018 ACM SIGIR International Conference
  on Theory of Information Retrieval}, ICTIR '18, pp.\  27–34, New York, NY,
  USA, 2018. Association for Computing Machinery.
\newblock ISBN 9781450356565.
\newblock \doi{10.1145/3234944.3234956}.
\newblock URL \url{https://doi.org/10.1145/3234944.3234956}.

\bibitem[Kejriwal et~al.(2017)Kejriwal, Ding, Shao, Kumar, and
  Szekely]{kejriwal2017flagit}
Kejriwal, M., Ding, J., Shao, R., Kumar, A., and Szekely, P.
\newblock Flagit: A system for minimally supervised human trafficking indicator
  mining.
\newblock \emph{arXiv preprint arXiv:1712.03086}, 2017.

\bibitem[Kumar(2020)]{kumar_how_2020}
Kumar, D.~V.
\newblock How to {Implement} {Convolutional} {Autoencoder} in {PyTorch} with
  {CUDA}, July 2020.

\bibitem[Li \& She(2017)Li and She]{10.1145/3126686.3126774}
Li, X. and She, J.
\newblock Relational variational autoencoder for link prediction with
  multimedia data.
\newblock In \emph{Proceedings of the on Thematic Workshops of ACM Multimedia
  2017}, Thematic Workshops '17, pp.\  93–100, New York, NY, USA, 2017.
  Association for Computing Machinery.
\newblock ISBN 9781450354165.
\newblock \doi{10.1145/3126686.3126774}.
\newblock URL \url{https://doi.org/10.1145/3126686.3126774}.

\bibitem[Radford et~al.(2021)Radford, Kim, Hallacy, Ramesh, Goh, Agarwal,
  Sastry, Askell, Mishkin, Clark, et~al.]{radford2021learning}
Radford, A., Kim, J.~W., Hallacy, C., Ramesh, A., Goh, G., Agarwal, S., Sastry,
  G., Askell, A., Mishkin, P., Clark, J., et~al.
\newblock Learning transferable visual models from natural language
  supervision.
\newblock In \emph{International Conference on Machine Learning}, pp.\
  8748--8763. PMLR, 2021.

\bibitem[Shi et~al.(2015)Shi, Chen, Wang, Yeung, Wong, and
  Woo]{shi2015convolutional}
Shi, X., Chen, Z., Wang, H., Yeung, D.-Y., Wong, W.-K., and Woo, W.-c.
\newblock Convolutional lstm network: A machine learning approach for
  precipitation nowcasting.
\newblock \emph{Advances in neural information processing systems}, 28, 2015.

\bibitem[Spotlight(2021)]{thorn}
Spotlight.
\newblock Thorn, 2021.
\newblock URL \url{https://www.thorn.org/spotlight/}.

\bibitem[Stickle \& Felson(2020)Stickle and Felson]{stickle2020crime}
Stickle, B. and Felson, M.
\newblock Crime rates in a pandemic: The largest criminological experiment in
  history.
\newblock \emph{American Journal of Criminal Justice}, 45\penalty0
  (4):\penalty0 525--536, 2020.

\bibitem[Sultana et~al.(2020)Sultana, Sufian, and Dutta]{sultana_review_2020}
Sultana, F., Sufian, A., and Dutta, P.
\newblock A {Review} of {Object} {Detection} {Models} based on {Convolutional}
  {Neural} {Network}.
\newblock \emph{arXiv:1905.01614 [cs]}, 1157:\penalty0 1--16, 2020.
\newblock \doi{10.1007/978-981-15-4288-6\_1}.
\newblock URL \url{http://arxiv.org/abs/1905.01614}.
\newblock arXiv: 1905.01614.

\bibitem[Tong et~al.(2017)Tong, Zadeh, Jones, and Morency]{tong2017combating}
Tong, E., Zadeh, A., Jones, C., and Morency, L.-P.
\newblock Combating human trafficking with deep multimodal models.
\newblock \emph{arXiv preprint arXiv:1705.02735}, 2017.

\bibitem[Wetzel(2017)]{wetzel_unsupervised_2017}
Wetzel, S.~J.
\newblock Unsupervised learning of phase transitions: {From} principal
  component analysis to variational autoencoders.
\newblock \emph{Physical Review E}, 96\penalty0 (2):\penalty0 022140, August
  2017.
\newblock \doi{10.1103/PhysRevE.96.022140}.
\newblock URL \url{https://link.aps.org/doi/10.1103/PhysRevE.96.022140}.

\bibitem[Yan et~al.(2020)Yan, Wang, Yang, Xu, and
  Huang]{10.1145/3388440.3412458}
Yan, C., Wang, S., Yang, J., Xu, T., and Huang, J.
\newblock Re-balancing variational autoencoder loss for molecule sequence
  generation.
\newblock In \emph{Proceedings of the 11th ACM International Conference on
  Bioinformatics, Computational Biology and Health Informatics}, BCB '20, New
  York, NY, USA, 2020. Association for Computing Machinery.
\newblock ISBN 9781450379649.
\newblock \doi{10.1145/3388440.3412458}.
\newblock URL \url{https://doi.org/10.1145/3388440.3412458}.

\bibitem[Zhang et~al.(2019)Zhang, Tong, Xu, and Maciejewski]{zhang_graph_2019}
Zhang, S., Tong, H., Xu, J., and Maciejewski, R.
\newblock Graph convolutional networks: a comprehensive review.
\newblock \emph{Computational Social Networks}, 6\penalty0 (1):\penalty0 11,
  November 2019.
\newblock ISSN 2197-4314.
\newblock \doi{10.1186/s40649-019-0069-y}.
\newblock URL \url{https://doi.org/10.1186/s40649-019-0069-y}.

\bibitem[Zhao et~al.(2021)Zhao, Hu, Dong, Huang, Weng, and
  Zhang]{zhao2021combination}
Zhao, J., Hu, L., Dong, Y., Huang, L., Weng, S., and Zhang, D.
\newblock A combination method of stacked autoencoder and 3d deep residual
  network for hyperspectral image classification.
\newblock \emph{International Journal of Applied Earth Observation and
  Geoinformation}, 102:\penalty0 102459, 2021.

\bibitem[Zheng et~al.(2021)Zheng, Sadhu, and Nevatia]{zheng2021improving}
Zheng, Z., Sadhu, A., and Nevatia, R.
\newblock Improving object detection and attribute recognition by feature
  entanglement reduction.
\newblock In \emph{2021 IEEE International Conference on Image Processing
  (ICIP)}, pp.\  2214--2218. IEEE, 2021.

\bibitem[Zhou et~al.(2020)Zhou, Wu, Lei, Hwang, and Yu]{zhou2020salient}
Zhou, W., Wu, J., Lei, J., Hwang, J.-N., and Yu, L.
\newblock Salient object detection in stereoscopic 3d images using a deep
  convolutional residual autoencoder.
\newblock \emph{IEEE Transactions on Multimedia}, 23:\penalty0 3388--3399,
  2020.

\end{thebibliography}
\bibliographystyle{icml2022}


\end{document}